%% file: main.tex
\documentclass[11pt]{article}

\usepackage{acl}

\usepackage{times}
\usepackage{latexsym}

\usepackage[T1]{fontenc}

\usepackage[utf8]{inputenc}

\usepackage{microtype}

\usepackage{inconsolata}

\usepackage{graphicx}

\usepackage{makecell}

\usepackage{booktabs}

\usepackage{float}

\usepackage{fvextra}

\usepackage{tabularx}

\title{ADRD-Bench: An LLM Benchmark for Alzheimer’s Disease and Related Dementias}

\author{
  \textbf{Guangxin Zhao\textsuperscript{1}}
  \quad
  \textbf{Jiahao Zheng\textsuperscript{2}}
  \quad
  \textbf{Malaz Boustani\textsuperscript{3}}
  \quad
  \textbf{Jarek Nabrzyski\textsuperscript{2}}
  \\
  \textbf{Yiyu Shi\textsuperscript{2}}
  \quad
  \textbf{Meng Jiang\textsuperscript{2}}
  \quad
  \textbf{Zhi Zheng\textsuperscript{1}}
  \\
  \\
  \textsuperscript{1}Electrical Engineering, University of Notre Dame
  \\
  \textsuperscript{2}Computer Science and Engineering, University of Notre Dame
  \\
  \textsuperscript{3}School of Medicine, Indiana University
  \\
  \small{
    \texttt{gzhao23@nd.edu},
    \texttt{jzheng7@nd.edu},
    \texttt{mboustan@iu.edu},
    \texttt{naber@nd.edu},
    \texttt{yshi4@nd.edu},
    \texttt{mjiang2@nd.edu},
    \texttt{zzheng3@nd.edu}
  }
}

\begin{document}
\maketitle
\begin{abstract}
Large language models (LLMs) have shown great potential for healthcare applications. However, existing evaluation benchmarks provide minimal coverage of Alzheimer’s Disease and Related Dementias (ADRD). To address this gap, we introduce ADRD-Bench, a preliminary ADRD-specific LLM benchmark. ADRD-Bench has two components: 1) ADRD Unified QA, a synthesis of 1,438 questions consolidated from seven established medical benchmarks, providing a unified assessment of clinical knowledge; and 2) ADRD Caregiving QA, a novel set of 149 questions derived from a nationally adopted, large clinical trials supported brain health management program, mitigating the lack of practical caregiving context in existing benchmarks. We evaluated 36 state-of-the-art LLMs on the proposed ADRD-Bench. Results showed that the accuracy of open-weight general models, open-weight medical models, and frontier closed-source general models ranged from 0.63 to 0.93 (mean: 0.77; std: 0.09), 0.47 to 0.93 (mean: 0.81; std: 0.14), and 0.83 to 0.93 (mean: 0.90; std: 0.03), respectively. While top-tier models achieved high accuracies (>0.9), case studies revealed inconsistent reasoning quality and stability, highlighting a critical need for domain-specific improvement to enhance LLMs’ knowledge and reasoning grounded in daily caregiving data.
\end{abstract}

\input{Content/1-Introduction}
\input{Content/3-ADRD_Benchmark_Construction}
\input{Content/4-Result_and_Analyses}
\input{Content/5-Conclusion}

\input{Content/Limitations}
\input{Content/Ethical_Considerations}



\bibliography{custom}

\appendix
\renewcommand{\thetable}{\thesection.\arabic{table}}
\setcounter{table}{0}

\input{Content/Appendix}

\end{document}

%% file: Content/1-Introduction.tex
\section{Introduction}
\label{sec:intro}

Large language models (LLMs) have achieved remarkable progress in healthcare applications. They have been used for assisting clinical decision support, providing medical education~\citep{zhou_survey_2023}, improving diagnostic reasoning, and enhancing patient care~\citep{maity_large_2025}. Models such as the GPT series and PaLM variants have achieved encouraging performance on medical examinations and specialty-specific tasks~\citep{kung_performance_2023, nori_capabilities_2023, chowdhery_palm_2023, singhal_large_2023}. However, using LLMs in healthcare carries substantial risks due to hallucinations~\citep{tang_evaluating_2023}, which can be dangerous in healthcare that requires high precision and rigor~\citep{pal_open_medical_llm_leaderboard_2024}. For example, incorrect drug dosages or misdiagnoses provided by LLMs may jeopardize patient safety~\citep{kim_medical_2025}. Therefore, rigorous domain-specific evaluation is a critical step before real-world clinical application~\cite{karabacak_embracing_2023,singhal_large_2023,zhou_survey_2023}.

Existing medical LLM benchmarks provide important foundations for evaluating general clinical knowledge. Broad benchmarks, such as MMLU, assess medical knowledge as parts of a wider multitask evaluation~\citep{hendrycks_measuring_2020}, while medical benchmarks, such as PubMedQA, HEAD-QA, MedQA, MedMCQA, and MedBullets, have been used to evaluate clinical knowledge recall and exam-style reasoning~\citep{jin_pubmedqa_2019,vilares_head-qa_2019,jin_what_2021,pal_medmcqa_2022,chen_benchmarking_2025}. More recent resources, including MultiMedQA, HealthBench, MEDEC, and MedHallu, further broaden evaluation toward consumer health questions, open-ended clinical generation, medical error detection, and hallucination detection~\citep{singhal_large_2023,arora_healthbench_2025,ben_abacha_medec_2025,pandit_medhallu_2025}. Despite these gains, high scores on structured, exam-style questions do not necessarily translate to real-world clinical competence, robustness, or trustworthiness~\citep{singhal_large_2023,kim_limitations_2025}. Furthermore, current benchmarks provide quite sparse and fragmented coverage of Alzheimer’s Disease and Related Dementias (ADRD).

Over 57 million people worldwide are living with ADRD as of 2021, and nearly 10 million new cases are diagnosed each year thereafter~\citep{who_dementia_2025}. Managing ADRD is not only a medical but also a socioeconomic challenge, as it heavily involves caregiving, long-term support, and complex decision-making by families~\citep{alzheimers_association_facts_2024}. LLMs are promising in serving ADRD care due to their capability of communication in natural language and cognitive support~\citep{treder_introduction_2024}. For example, with proper design, LLM-driven content may boost patients' cognitive abilities and enrich interactions, while improving diagnostic strategies and patient-caregiver communication~\citep{treder_introduction_2024}. Like any other medical device, LLMs for ADRD care should be evaluated for technical performance before planning for real-world deployment~\citep{who_ai_ethics_2024}. 

Prior ADRD-related NLP research has focused primarily on detection and diagnosis, such as screening cognitive impairment or classifying dementia from speech and text~\citep{agbavor_predicting_2022,peled-cohen_systematic_2025}. A recent systematic review reported that approximately 56\% of ADRD-related NLP studies focused on cognitive impairment detection, whereas only about 6\% addressed caregiver support~\citep{peled-cohen_systematic_2025}. Although emerging systems such as ADQueryAid and the Caregiving Language Model (CaLM) have begun to explore caregiver-oriented dialogue and dementia-care-specific language modeling~\citep{hasan_empowering_2024,parmanto_reliable_2024}, rigorous LLM benchmarks for ADRD care remain underdeveloped.

To the best of our knowledge, there is no existing ADRD-specific LLM benchmark for this purpose. Some existing medical benchmarks include ADRD-related items; however, these items constitute only a tiny proportion of the total content, mostly around 1\% (see Section~\ref{sec:adrd_unified_qa}). In addition, there are almost no caregiving types of testing in existing datasets. While LLMs are often targeted to provide support for caregivers, missing the caregiving type leaves a huge gap for real-world applications.

To mitigate this gap, we introduce ADRD-Bench, a preliminary ADRD-specific LLM benchmark. This benchmark has two parts (ADRD Unified QA set and ADRD Caregiving QA set) constructed from two complementary sources. The ADRD Unified QA set includes 1,438 ADRD-related items extracted from seven existing benchmark datasets. By aggregating these scattered items into a single set, we create a concentrated evaluation of how well current models handle ADRD knowledge that was previously only sparsely represented. The ADRD Caregiving QA set includes 149 new questions derived from the well-validated Aging Brain Care (ABC) program, which a leading evidence-based dementia care models~\citep{alz_association_dementia_2024,alder2024indiana,lamantia_aging_2015} in the U.S. From de-identified ABC caregiver education materials, we crafted question-answer pairs that reflect realistic caregiving content, including daily caring situations (e.g., patient aggressive behavior management at home). Although modest in size, the ADRD Caregiving QA set presents the capability to disclose performance disparities among SOTA models in practical ADRD caregiving scenarios.

In summary, our work presents the first benchmark tailored to ADRD tasks, which brings together a wide range of ADRD caring needs (e.g., factual Q\&A, clinical reasoning, caregiver guidance). This resource enables a focused technical assessment of LLM capabilities in a domain where factual accuracy and empathetic reasoning are paramount. We hope that evaluating LLMs on ADRD-specific tasks will help identify their current capabilities and limitations as well as drive the development of safer, more reliable medical AI for the science of ADRD care.

%% file: Content/3-ADRD_Benchmark_Construction.tex
\section{ADRD Benchmark Construction}
\label{sec:construction}

\subsection{ADRD Unified QA}
\label{sec:adrd_unified_qa}

To identify ADRD-related questions from existing benchmarks, ADRD keywords, including "Alzheimer's", "Dementia", "ADRD", "AD", "cognitive decline", "cholinesterase inhibitors", "donepezil", "rivastigmine", and "galantamine", were used to search through 12 popular public benchmark datasets (full results in Appendix Table~\ref{tab:adrd-question-statistics}.). Except for MEDIQA, the proportion of ADRD-related questions in each benchmark is below 3\%, and most are around 1\%. The proportion of MEDIQA is 13.32\%; however, since it contains 383 questions, 13.32\% leads to 51 questions only. Across all datasets, the total number of ADRD-related questions accounts for 0.68\% of total non-generation task questions and 0.40\% of generation task questions, respectively. Therefore, existing medical benchmarks contain only a tiny proportion of ADRD-relevant content, indicating a substantial gap and leaving significant room for exploration. 

To build the ADRD Unified QA set, we consolidated ADRD questions from seven non-generation task benchmarks, including PubMedQA, HEAD-QA, MedBullets, MedMCQA, MedQA, MEDEC, and MedHallu, which are the leading healthcare benchmarks to test SOTA LLMs~\cite{kim_questioning_2025,singhal_large_2023}. MEDCALC-BENCH~\cite{khandekar_medcalc-bench_2024} is excluded since it focuses on calculation and arithmetic reasoning instead of ADRD-related knowledge and reasoning. In addition, ADRD Unified QA does not include generation tasks, as existing evaluation metrics are insufficient to reliably and consistently assess the quality of LLM responses~\cite{singhal_large_2023}. Thus, the ADRD Unified QA consists of two primary question types: 1) multiple-choice questions to evaluate models' knowledge of ADRD-related concepts, which were from PubMedQA, HEAD-QA, MedBullets, MedMCQA, and MedQA; and 2) error detection questions, assessing models' ability to identify hallucinations and factual errors present in textual inputs, which were from MEDEC and MedHallu.

To characterize the content of ADRD Unified QA, care-context categorization of all 1,438 questions was needed. Guided by literature on dementia care \citep{lin_evidence_2025, tong_navigating_2026, alzheimers_association_facts_2025} and working with a nationally certified clinician, five categories are identified to be important for LLM applications: Caregiving, Clinical Nursing, Memory Care, General Non-Memory Medical or Surgical Care, and Others. Using an LLM-assisted procedure (see Appendix~\ref{unified_qa_taxonomy} for detailed definitions, prompt, and expert verification), we identified 6, 4, 468, 502, and 458 questions in each category, respectively. This distribution indicated that existing benchmark questions are predominantly concentrated in specialty clinical care, general medical care, and broader peripheral domains, whereas questions directly addressing practical caregiving or clinical nursing are extremely rare. This finding further motivated the construction of ADRD Caregiving QA, which targets realistic daily caregiving scenarios. 

\subsection{ADRD Caregiving QA}
Managing ADRD in real-world settings extends far beyond diagnosis and pharmacological treatment. It largely revolves around day-to-day caregiving challenges such as behavioral management, communication with patients, safety supervision, and caregiver decision-making in home and community settings. To fill the gap that almost no daily caregiving questions present in existing benchmarks (see Section~\ref{sec:adrd_unified_qa}), we constructed the ADRD Caregiving QA set based on materials from the Aging Brain Care (ABC) program. The ABC program is a comprehensive, nationally adopted, evidence-based model of ADRD care that has been continuously developed, evaluated, and implemented for nearly two decades~\cite{alz_association_dementia_2024,alder2024indiana,lamantia_aging_2015}. More details of ABC program are in Appendix~\ref{abc_program}.
The ABC program naturally serves as a solid foundation to extract benchmark questions. The ADRD Caregiving QA set covers essential dimensions of ADRD caregiving, such as understanding the condition, effective communication with patients, and behavioral management strategies. 

We followed a structured abstraction process to transform recurring caregiving situations and recommended care strategies from the ABC program's well-validated educational content and caregiver handouts into evaluable question-answer pairs. The resulting QA set included 120 polar (True/False) questions and 29 multiple-choice questions. For True/False questions, statements were adapted directly from the source materials, with false statements constructed by modifying critical words to create incorrect assertions. For multiple-choice questions, existing cases and descriptions in the materials were reworded into questions and answer options; when additional distractors were needed but were not available in the ABC materials, Gemini-3-Pro was used for generation. To ensure data quality, all constructed QAs were reviewed and refined for accuracy, clarity, and appropriateness by a senior clinician who served as a lead designer of the ABC program and has been involved in its development and refinement for two decades. Since the ABC materials were already well validated for the validity of the benchmark QA, this human examination focused on ensuring that the QA texts were aligned with the ABC documents. The clinician paid particular attention to questions requiring synthesis across multiple materials, rewording, or newly added wording, and revised them as needed.

By integrating the ADRD Caregiving QA set with the ADRD Unified QA set, the ADRD-Bench enables a more comprehensive evaluation of LLMs that spans both clinical knowledge and practical caregiving support for ADRD. 

\subsection{Quantitative and Qualitative Evaluations}
To validate SOTA LLMs on the proposed ADRD-Bench, we tested 36 models as listed in Table~\ref{tab:1}, including both open-weight LLMs and closed-source LLMs. The open-weight LLMs included general and medical LLMs, varying in size from 3.8B to 235B parameters. This evaluation on a diverse set of models with different parameter scales and deployment settings established not only a comprehensive understanding of the models' performance but also the overall difficulty level of the benchmark across various model settings. Details of the inference hardware are provided in Appendix~\ref{hardware}.

For quantitative evaluation, given that all tasks in ADRD-Bench were formulated as discrete-choice questions, including multiple-choice, True/False, and error sentence detection formats, we adopted exact match accuracy as the primary metric. A model's prediction was considered correct if and only if the selected option exactly matched the ground-truth label. For each model, we reported accuracies separately for each set of the benchmark and the overall accuracy aggregated across all questions.

For qualitative analysis, we focused on the newly proposed ADRD Caregiving QA set to better understand model failures in daily caregiving scenarios, which have not been covered in the literature. Specifically, we selected questions on which LLMs frequently produced incorrect answers. The qualitative analyses were intended to contextualize the findings and highlight practical limitations of current LLMs in daily caregiving scenarios.

%% file: Content/4-Result_and_Analyses.tex
\section{Results and Analyses}
\label{sec:result}

Table ~\ref{tab:1} shows the accuracy of 36 LLMs on each subset of ADRD-Bench. LLMs are grouped into open-weight general, open-weight medical, and closed-source general categories, with model size indicated where applicable. Boldface indicates the best performance for each subset, while underlining denotes the second best.

\begin{table*}[t]
\centering
\scriptsize
\setlength{\tabcolsep}{4pt}
\renewcommand{\arraystretch}{1.12}
\caption{Accuracy of LLMs on ADRD-Bench.}
\label{tab:1}

\begin{tabular}{l | c | cccccccc | ccc}
\hline
 &  & \multicolumn{8}{c|}{\textbf{ADRD Unified QA}} & \multicolumn{3}{c}{\textbf{ADRD Caregiving QA}} \\
\cline{3-13}
\multicolumn{1}{c|}{\textbf{Model Name}} & \makecell{\textbf{Model}\\\textbf{Size}}
& \makecell{Med \\\ MCQA} & \makecell{Med \\\ QA} & \makecell{HEAD- \\\ QA} & \makecell{MED \\\ EC} & \makecell{Med \\\ Bullets} & \makecell{Med \\\ Hallu} & \makecell{PubMed \\\ QA} & Score
& \makecell{True/ \\\ False} & \makecell{Multiple \\\ Choice} & Score \\
 &  & 1,193 (size) & 119 & 43 & 37 & 18 & 16 & 12 & 1,438
& 120 & 29 & 149 \\
\hline

\multicolumn{13}{l}{\textbf{Open-Weight General Model}} \\
\hline
Phi-3-mini-128k-instruct & 3.8B 
& 0.6438 & 0.6134  & 0.7907 & 0.0811 & 0.5000 & 0.7500 & \textbf{0.8333} & 0.6321 & 0.8583 & 0.8276 & 0.8523 \\
Qwen3-4B-Instruct-2507 & 4B
& 0.7117 & 0.6387 & 0.8140 & 0.1622 & 0.5556 & 0.8750 & \textbf{0.8333} & 0.6954 & 0.9000 & 0.7586 & 0.8725 \\
Falcon3-7B-Instruct & 7.5B
& 0.6857 & 0.5714 & 0.7209 & 0.2432 & 0.5556 & 0.6250 & \textbf{0.8333} & 0.6648 & 0.8583 & 0.7241 & 0.8322 \\
Qwen2.5-7B-Instruct & 7.6B
& 0.7318 & 0.5882 & 0.7209 & 0.3514 & 0.5556 & 0.8125 & \underline{0.7500} & 0.7086 & 0.8917 & 0.7241 & 0.8591 \\
Meta-Llama-3-8B-Instruct & 8B
& 0.9246 & 0.6975 & 0.7209 & 0.1622 & 0.6667 & 0.6875 & \underline{0.7500} & 0.8727 & 0.8417 & 0.6897 & 0.8121 \\
Llama-3.1-8B-Instruct & 8B
& 0.8759 & 0.6975 & 0.7674 & 0.0811 & 0.6667 & 0.7500 & \textbf{0.8333} & 0.8331 & 0.8583 & 0.7586 & 0.8389 \\
Yi-1.5-9B-Chat & 9B
& 0.6178 & 0.5378 & 0.7442 & 0.1081 & 0.3889 & 0.8125 & \textbf{0.8333} & 0.6029 & 0.8750 & 0.7241 & 0.8456 \\
gemma-2-9b-it & 9.2B
& 0.7393 & 0.5882 & 0.7442 & 0.2703 & 0.6667 & 0.8750 & \underline{0.7500} & 0.7156 & 0.8667 & 0.7931 & 0.8523 \\
gemma-3-12b-it & 12.2B
& 0.7410 & 0.6218 & 0.7907 & 0.1892 & 0.6667 & 0.8125 & \underline{0.7500} & 0.7184 & 0.9333 & 0.7931 & 0.9060 \\
Phi-3-medium-4k-instruct & 14B
& 0.7636 & 0.7059 & 0.8140 & 0.2703 & 0.6667 & 0.8125 & \textbf{0.8333} & 0.7476 & 0.9250 & 0.7586 & 0.8926 \\
Qwen2.5-14B-Instruct & 14.8B
& 0.7795 & 0.6387 & 0.7907 & 0.2432 & 0.5556 & 0.8125 & \underline{0.7500} & 0.7517 & 0.8333 & 0.7241 & 0.8121 \\
Qwen3-30B-A3B-Instruct-2507 & 30B
& 0.8466 & 0.7479 & 0.8372 & 0.4595 & 0.8333 & 0.8125 & \underline{0.7500} & 0.8268 & 0.9083 & 0.7586 & 0.8792 \\
Llama-3.1-70B-Instruct & 70B
& \underline{0.9505} & 0.9412 & 0.8372 & 0.4324 & 0.7222 & \textbf{1.0000} & \textbf{0.8333} & \textbf{0.9298} & \underline{0.9583} & 0.7931 & 0.9262 \\
Qwen2.5-72B-Instruct & 70B
& 0.8583 & 0.7479 & 0.8605 & 0.2703 & 0.6667 & 0.8125 & \underline{0.7500} & 0.8303 & 0.9167 & 0.8276 & 0.8993 \\
Qwen3-235B-A22B-Instruct-2507 & 235B
& 0.8910 & 0.8908 & 0.8837 & 0.4595 & 0.7222 & \underline{0.9375} & \underline{0.7500} & 0.8769 & 0.9417 & \underline{0.8966} & 0.9329 \\
\hline

\multicolumn{13}{l}{\textbf{Open-Weight Medical Model}} \\
\hline
medgemma-4b-it & 4B
& 0.7360 & 0.5966 & 0.7907 & 0.2162 & 0.6111 & 0.6250 & 0.6667 & 0.7093 & 0.8667 & 0.5862 & 0.8121 \\
Hulu-Med-4B & 4B
& 0.8684 & 0.8067 & 0.8140 & 0.4054 & 0.5000 & 0.7500 & \textbf{0.8333} & 0.8435 & 0.8500 & 0.8276 & 0.8456 \\
Lingshu-7B & 7B
& 0.7016 & 0.6975 & 0.8140 & 0.2162 & 0.6111 & 0.7500 & \underline{0.7500} & 0.6919 & 0.8583 & 0.6897 & 0.8255 \\
Hulu-Med-7B & 7B
& 0.9095 & 0.8235 & 0.7907 & 0.2973 & 0.7222 & 0.8125 & \textbf{0.8333} & 0.8790 & 0.9083 & 0.7931 & 0.8859 \\
BioMistral-7B & 7B
& 0.4602 & 0.4286 & 0.6512 & 0.2973 & 0.3889 & 0.7500 & 0.5000 & 0.4618 & 0.6083 & 0.4483 & 0.5772 \\
Hulu-Med-14B & 14B
& 0.9279 & 0.9328 & 0.8372 & 0.3243 & \underline{0.8889} & 0.8125 & \textbf{0.8333} & 0.9075 & 0.9417 & 0.7586 & 0.9060 \\
Hulu-Med-32B & 32B
& 0.9396 & 0.9664 & 0.8837 & 0.2973 & 0.7222 & 0.8750 & \textbf{0.8333} & 0.9193 & \underline{0.9583} & 0.7931 & 0.9296 \\
Lingshu-32B & 32B
& 0.8248 & 0.9160 & 0.8140 & 0.4054 & \textbf{0.9444} & 0.8125 & \textbf{0.8333} & 0.8227 & \textbf{0.9750} & 0.8621 & \underline{0.9530} \\
Llama3-OpenBioLLM-70B & 70B
& 0.9464 & 0.8992 & 0.8140 & 0.3784 & 0.7778 & 0.7500 & \underline{0.7500} & 0.9179 & 0.9250 & 0.7931 & 0.8993 \\
ClinicalCamel-70B & 70B
& 0.6521 & 0.5714 & 0.7442 & 0.3514 & 0.3889 & \underline{0.9375} & 0.5833 & 0.6398 & 0.8750 & 0.7586 & 0.8523 \\
Med42-70B & 70B
& 0.9053 & 0.8908 & 0.8605 & 0.2973 & 0.7778 & \underline{0.9375} & \textbf{0.8333} & 0.8853 & 0.9167 & 0.7586 & 0.8859 \\
Meditron3-70B & 70B
& 0.9480 & 0.9160 & \underline{0.9070} & 0.5135 & 0.6667 & \underline{0.9375} & \textbf{0.8333} & \underline{0.9284} & 0.9417 & 0.7931 & 0.9128 \\
MMed-Llama3.1-70B & 70B
& \textbf{0.9564} & 0.9328 & \textbf{0.9302} & 0.0270 & 0.5000 & 0.6250 & \underline{0.7500} & 0.9186 & 0.9250 & 0.7586 & 0.8926 \\
\hline

\multicolumn{13}{l}{\textbf{Closed-Source General Model}} \\
\hline
gemini-2.5-pro-review & --
& 0.9212 & 0.9580 & \textbf{0.9302} & 0.4865 & \underline{0.8889} & 0.8750 & \textbf{0.8333} & 0.9117 & 0.9333 & 0.7931 & 0.9060 \\
gemini-3.1-pro-review & --
& 0.9246 & \textbf{0.9832} & \textbf{0.9302} & 0.5135 & 0.7778 & \underline{0.9375} & \textbf{0.8333} & 0.9166 & 0.9333 & \textbf{0.9655} & 0.9396 \\
qwen3-max-2025-09-23 & --
& 0.9028 & 0.9328 & \underline{0.9070} & 0.4054 & 0.7778 & \underline{0.9375} & \textbf{0.8333} & 0.8908 & 0.9417 & 0.8276 & 0.9195 \\
gpt-5.2-2025-12-11 & --
& 0.8969 & 0.8908 & \underline{0.9070} & \textbf{0.8649} & \underline{0.8889} & 0.8125 & \textbf{0.8333} & 0.8943 & 0.9333 & \textbf{0.9655} & 0.9396 \\
gpt-5.5-2026-04-23 & --
& 0.9380 & \underline{0.9748} & \textbf{0.9302} & 0.5135 & \underline{0.8889} & 0.8750 & \textbf{0.8333} & 0.9277 & \underline{0.9583} & \textbf{0.9655} & \textbf{0.9597} \\
claude-opus-4.5-20251101 & --
& 0.9204 & 0.9328 & \textbf{0.9302} & \underline{0.5946} & \underline{0.8889} & 0.8125 & \textbf{0.8333} & 0.9110 & 0.9333 & \underline{0.8966} & 0.9262 \\
claude-opus-4.7 & --
& 0.9313 & \textbf{0.9832} & \textbf{0.9302} & 0.4595 & \underline{0.8889} & \underline{0.9375} & \textbf{0.8333} & 0.9221 & 0.9500 & \textbf{0.9655} & \underline{0.9530} \\
grok-4.1-fast-non-reasoning & --
& 0.8315 & 0.8403 & 0.8605 & 0.3514 & 0.7222 & \underline{0.9375} & \textbf{0.8333} & 0.8206 & 0.9000 & 0.8276 & 0.8859 \\
\hline
\end{tabular}

\vspace{2pt}
\begin{flushleft}
\footnotesize Boldface: the best performance; Underlining: the second-best result. Within each QA set, all questions across subcategories were equally weighted; scores were computed over all questions.
\end{flushleft}
\end{table*}

\subsection{LLM Performance on ADRD Unified QA}
\begin{figure}[t]
    \centering
    \includegraphics[width=1\linewidth]{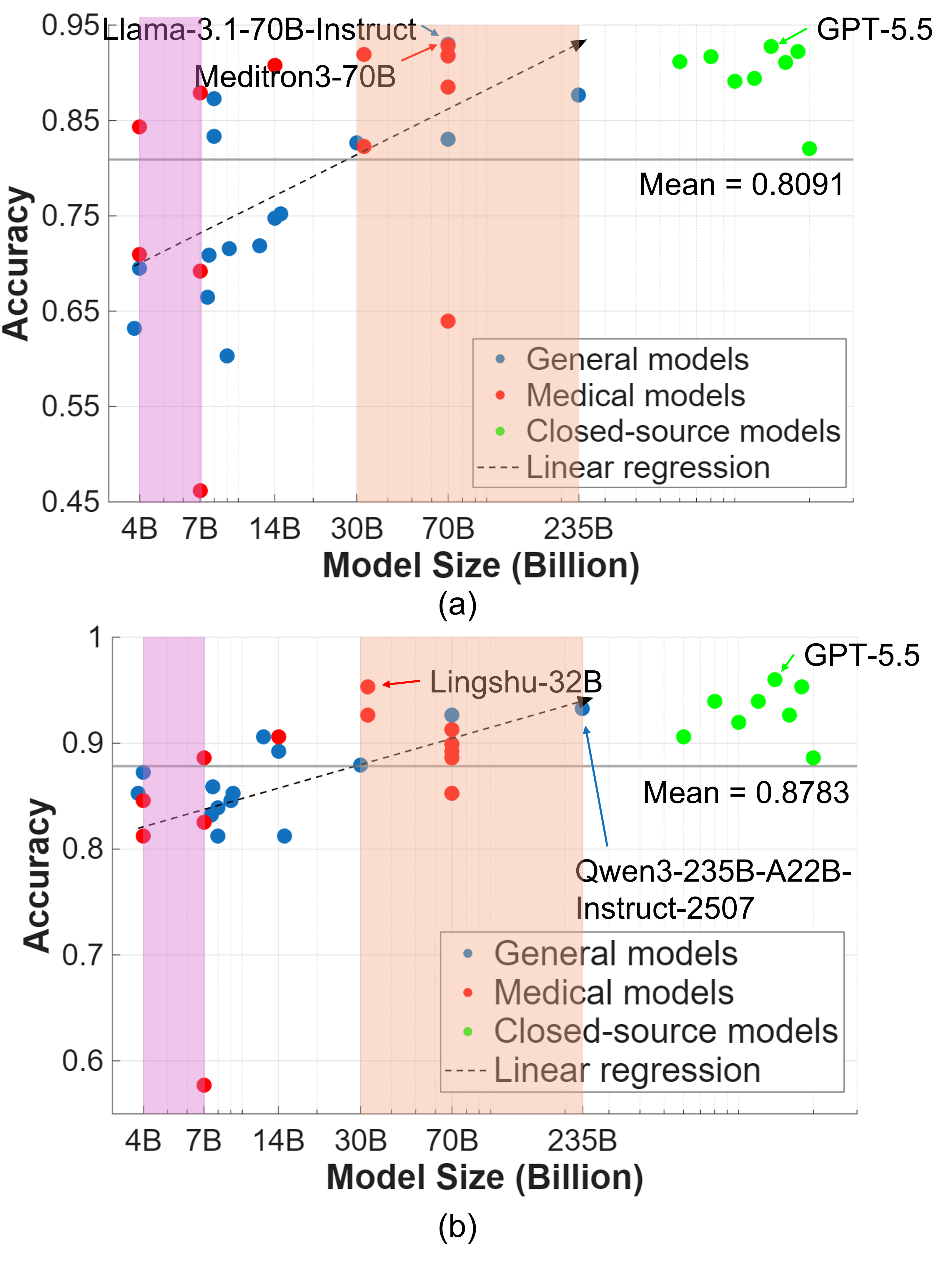}
    \caption{
    Model accuracy vs. model size on the (a) ADRD Unified QA and (b) ADRD Caregiving QA. Blue points: open-weight general models; Red points: open-weight medical models; Green points: closed-source general models. Dashed horizontal line: mean accuracy across all models.
    }
    \label{fig:1}
\end{figure}

To understand the association between model size and accuracy, statistical analyses were conducted. Figure~\ref{fig:1} (a) shows a significant, moderate-to-strong positive correlation (Pearson's $r = 0.5443$, $p = 0.0027$) between model size and accuracy on the ADRD Unified QA set. Most smaller models with 4--7B parameters achieved relatively low performance (approximately $0.45$--$0.75$), whereas large models (70B+ parameters) demonstrated substantially higher accuracy, mostly $\geq 0.85$. Moreover, most models with more than 30B parameters performed above the overall mean accuracy (0.8091). The best-performing model was Llama-3.1-70B-Instruct~\cite{meta_ai_llama-31-70b-instruct_2024}, which achieved an accuracy of $0.9298$.

When comparing different model categories, open-weight medical models (red points) generally outperformed open-weight general models (blue points) at comparable sizes. This highlighted the benefits of domain-specific fine-tuning for clinical question answering and medical error detection. However, an exception was observed for the top-performing model, Llama-3.1-70B-Instruct, which is a general model and slightly outperformed the strongest medical model, Meditron3-70B~\cite{openmeditron_meditron3-70b_2025} (0.9284). All closed-source general models achieved performance above the overall mean accuracy, with gpt-5.5-2026-04-23~\cite{openai_introducing_gpt55_2026} performing the best among them, reaching an accuracy of $0.9277$. Nevertheless, it was slightly surpassed by several open-weight models (e.g., Llama-3.1-70B-Instruct and Meditron3-70B) with significantly fewer parameters. In addition, some open-weight medical LLMs exhibited substantially lower accuracy than open-weight general models with similar parameter counts, indicating that their domain-specific fine-tuning might not include ADRD.

Overall, these results suggest that model size is an important contributor to performance on the ADRD Unified QA; however, it does not fully determine model effectiveness. Domain-specific fine-tuning may provide performance advantages in certain cases, but general models can still outperform specialized models, possibly due to stronger general reasoning capabilities and more robust language representations learned from large-scale and diverse pretraining corpora.

\subsection{LLM Performance on ADRD Caregiving QA}
On the ADRD Caregiving QA set (Figure~\ref{fig:1} (b)), the performance gap between smaller and larger models was narrower than that observed on the ADRD Unified QA set. This might be due to the fact that the questions in this benchmark focus more on daily living and are closer to ``common-sense'' reasoning, which is generally well covered by SOTA LLMs at various sizes. Nevertheless, an upward trend in accuracy with increasing model size remained evident. Pearson's correlation analysis revealed a significant, moderate-to-strong positive correlation between model size and accuracy ($r = 0.4947$, $p = 0.0075$). Most smaller models in the 4B--7B parameter range achieved accuracies between $0.80$ and $0.86$, all of which (except for the Hulu-Med-7B~\cite{jiang2025hulu}) fell below the overall mean accuracy of $0.8783$. In contrast, the majority of models with more than 14B parameters performed above the average accuracy.

Except grok-4-1-fast-non-reasoning~\cite{xai_grok_2025}, all other closed-source general models achieved accuracies above $0.9$, outperforming the majority of open-weight models. This observation suggests that closed-source general models currently exhibit stronger overall capabilities in ADRD daily caregiving tasks. gpt-5.5-2026-04-23 achieved the best performance among all evaluated models, with an accuracy of $0.9597$, slightly surpassing the best-performing open-weight medical model, Lingshu-32B~\cite{xu_lingshu_2025} ($0.9530$), as well as the best-performing open-weight general model, Qwen3-235B-A22B-Instruct-2507~\cite{yang_qwen3_2025} ($0.9329$).

Medical models did not demonstrate a clear performance advantage over general models. In a few cases, they underperformed general models of comparable size. This pattern suggested that answering practical daily caregiving questions relied more on the broad commonsense reasoning capabilities of LLM, rather than on specialized medical fine-tuning. Meanwhile, this observation might reflect the fact that current domain-specific fine-tuning primarily focuses on medical knowledge and incorporates relatively limited information about daily caregiving.

To examine whether the True/False format induces a systematic affirmation bias, we calculated class-conditional accuracies ($\mathrm{True}\!\rightarrow\!\mathrm{True}\ \mathrm{vs.}\ \mathrm{False}\!\rightarrow\!\mathrm{False}$). Results showed an average $\mathrm{True}\!\rightarrow\!\mathrm{True}$ accuracy of $93.99\% \pm 5.11\%$ STD and a $\mathrm{False}\!\rightarrow\!\mathrm{False}$ accuracy of $83.81\% \pm 17.10\%$ STD across all models. Despite the higher accuracy on true statements, the modest mean gap and the large variability in $\mathrm{False}\!\rightarrow\!\mathrm{False}$ accuracy did not provide strong evidence of a uniform systematic affirmation bias across models.

\subsection{Unified QA vs. Caregiving QA Correlations}
\begin{figure}
    \centering
    \includegraphics[width=1\linewidth]{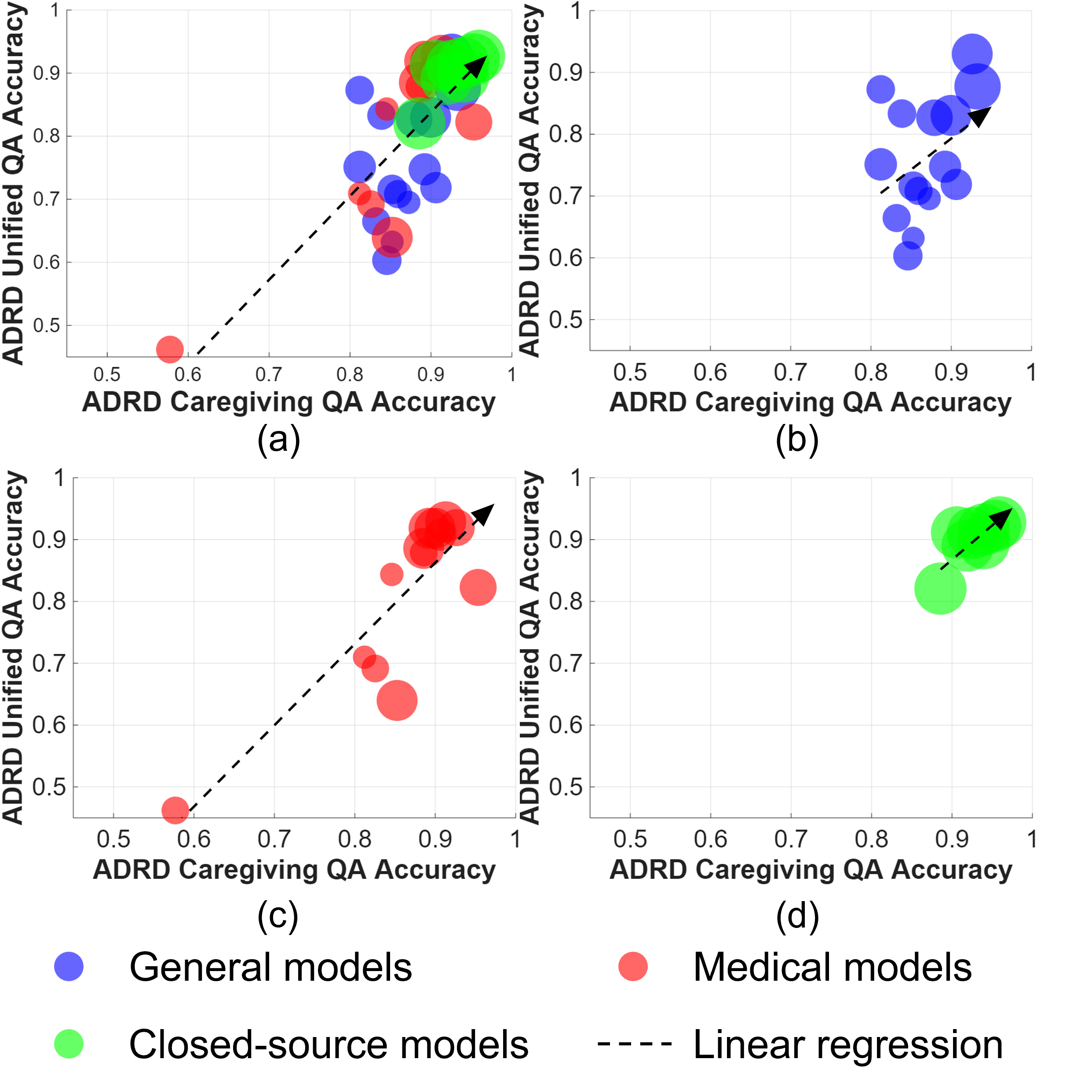}
    \caption{
    Correlation between accuracies on ADRD Caregiving QA and accuracies on ADRD Unified QA on (a) all models, (b) open-weight general models; (c) open-weight medical models; (d) closed-source general models, where larger points mean larger model parameters.
    }
    \label{fig:2}
\end{figure}

To understand how models' performance on clinical knowledge was related to that on caregiving, we analyzed the correlation between accuracies on ADRD Caregiving QA and accuracies on the ADRD Unified QA, as shown in Figure~\ref{fig:2}. Across all models, the Pearson’s correlation analysis revealed a significant and strong positive correlation between the two QA sets ($r = 0.7511$, $p < 0.0001$). This indicated that, at an aggregated level, models that performed well on clinical questions also tended to achieve a higher accuracy on the caregiving questions. However, by examining different model categories separately, open-weight medical models ($r = 0.8634$, $p = 0.0001$) and closed-source models ($r = 0.7987$, $p = 0.0174$) exhibited a significant and strong correlation. In contrast, open-weight general models ($r = 0.3961$, $p = 0.1439$) showed only weak-moderate, non-significant positive correlations.

 These results demonstrated that open-weight medical models, which incorporated specific medical fine-tuning, had more consistent performance across both clinical and caregiving questions than the open-weight general models. A possible cause of the non-significant correlations of the open-weight general models could be the relatively small portion of medical data in model training. Therefore, the models' performance depends more on common sense and social reasoning, which contribute to more socially appropriate responses, a key point in daily caregiving. This analysis also aligned with our earlier observation that, on ADRD Caregiving QA, open-weight general models achieved higher accuracy than open-weight medical models at comparable parameter scales. These findings also implied that, to develop healthcare LLMs that fit ADRD care needs, more training data on caregiving tasks should be included to balance the models' knowledge and reasoning on this aspect.

\subsection{Case Analysis}
To better understand LLM performance beyond numerical metrics, we examined the models' reasoning processes on ADRD Caregiving QA questions. Two representative cases are presented below to illustrate common failure patterns and model behaviors. Feedback from a clinical expert was incorporated to ensure that the analysis was grounded in real-world ADRD caregiving principles. In addition, within each model category (open-weight general, open-weight medical, and closed-source general), we selected two well-performing models for manual qualitative error taxonomy analysis. The models, error taxonomy, category definitions, frequencies, and percentages are provided in Appendix~\ref{tab:error_taxonomy_distribution}. 

\subsubsection{Case 1: True/False question}
\begin{figure*}
    \centering
    \includegraphics[width=1\linewidth]{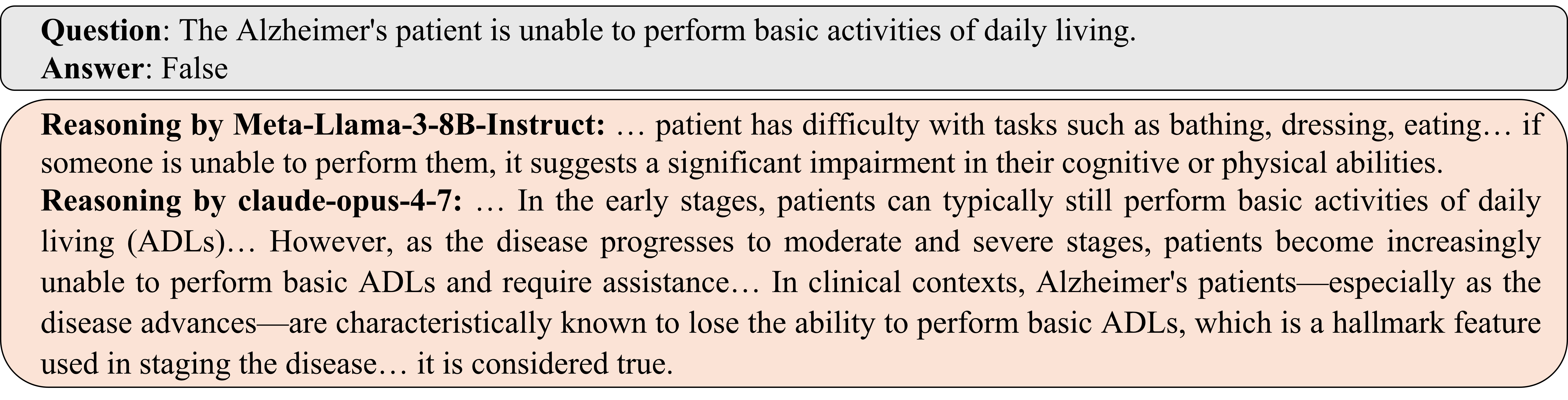}
    \caption{
    True/False question example of ADRD Caregiving QA that most LLMs answered incorrectly}
    \label{fig:3}
\end{figure*}

In Figure~\ref{fig:3}, the correct answer is ``False,'' as the statement overgeneralizes across all stages of Alzheimer's disease. In the early stages, individuals are often still able to perform most basic activities independently, and the loss of self-care ability typically occurs in moderate to severe ADRD stages. However, $83\%$ LLMs answered ``True,'' implying that all individuals with Alzheimer's disease are universally unable to manage daily living activities. This result reflected a common failure pattern of sweeping statements (e.g., overgeneralization).

From the LLMs' reasoning steps obtained via Zero-Shot Chain-of-Thought prompting shown in Figure~\ref{fig:3}, the output of claude-opus-4-7 demonstrated another important type of error. This model achieved an overall accuracy exceeding $90\%$ on the ADRD Caregiving QA and was able to distinguish between different stages of Alzheimer's disease and recognize that patients in the early stage generally retain the ability to perform activities of daily living. However, this LLM made its final judgment based on a hallmark of the advanced stage and used it to generalize a conclusion across the entire disease course, producing an incorrect answer. This suggested a tendency to rigidly focus on salient features, while overlooking conditional factors like disease progression. Complete prompts and outputs are in the Appendix~\ref{prompt_tf_cot} and~\ref{reasoning_tf}.

\subsubsection{Case 2: Multiple-Choice question}
\begin{figure*}[!t]
    \centering
    \includegraphics[width=1\linewidth]{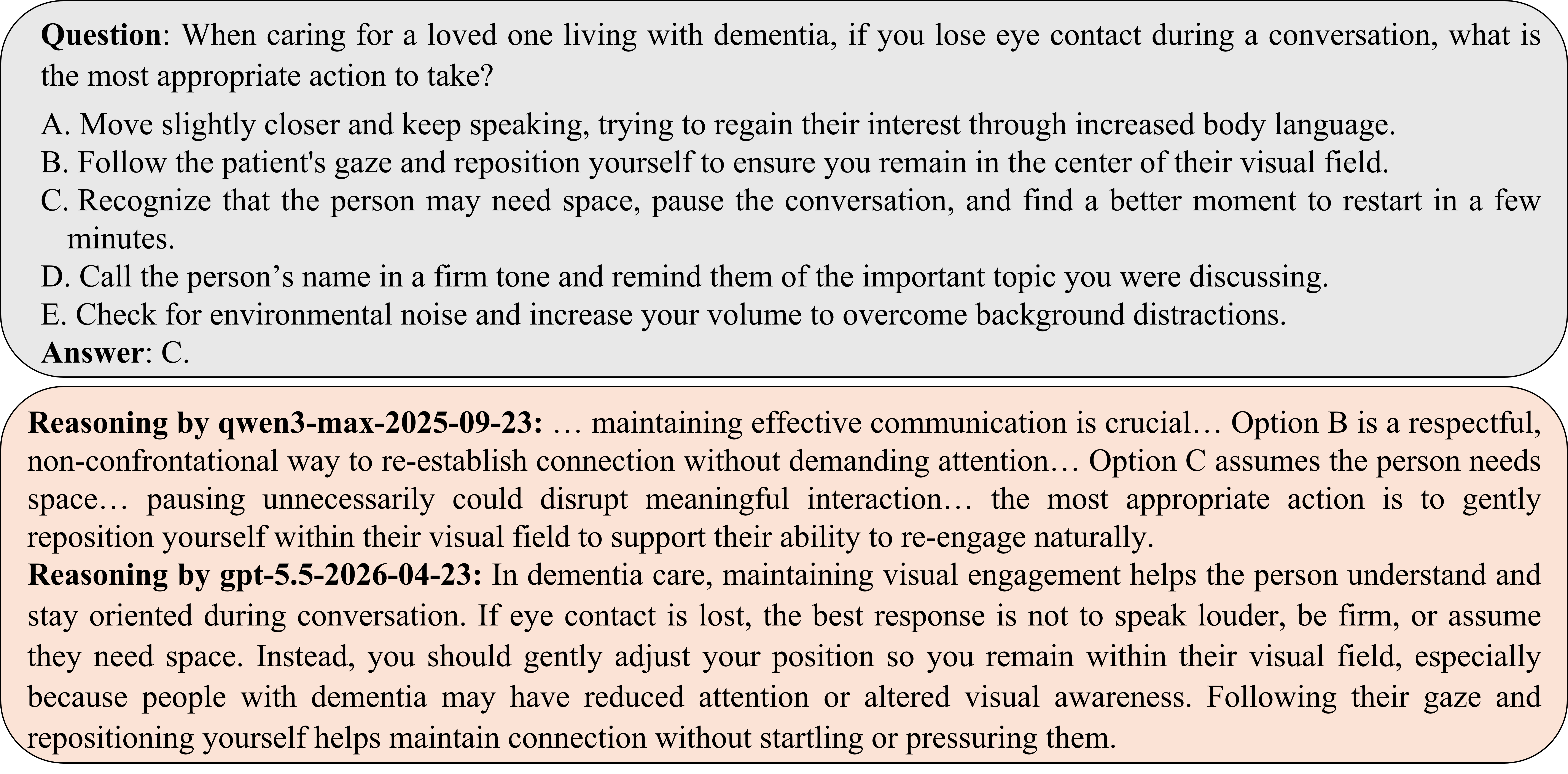}
    \caption{
    Multiple-choice example of ADRD Caregiving QA that most models answered incorrectly.
    }
    \label{fig:4}
\end{figure*}

Figure~\ref{fig:4} shows a multiple-choice question example. In patients living with ADRD, losing eye contact during a conversation often signals cognitive overload, fatigue, or a need to disengage, rather than a visual tracking problem. Thus, the recommended approach is to respect the individual’s need for space and reduce stimulation, rather than attempting to immediately re-engage them. However, $81\%$ LLMs chose ``B''. It reflects an LLM tendency to treat behavioral cues as attention problems to be corrected, prioritizing task or conversation continuity over respecting disengagement and emotion regulation.

Figure~\ref{fig:4} presents examples of LLMs’ key reasoning steps. The prompting and the entire outputs are in the Appendix~\ref{prompt_mc_cot} and~\ref{reasoning_mc}. Both qwen3-max-2025-09-23 and gpt-5.5-2026-04-23 achieved an overall accuracy of $> 90\%$ on the ADRD Caregiving QA. However, both failed to answer this question. The models could identify that a patient living with dementia might experience attentional, perceptual, or visual field challenges. Then, both subsequently treated this partial clinical insight as sufficient justification for active re-engagement, recommending the caregiver to reposition and thus re-enter the individual’s visual field. This reasoning reflects a form of misunderstanding of caring/medical
priority/focus. Rather than evaluating caregiving interaction through the loss of eye contact, the models focused on preserving conversational continuity. In doing so, they implicitly assumed that maintaining engagement is always desirable, failing to recognize that, in ADRD care, averted gaze is often a low-intensity disengagement or overload signal rather than a problem to be corrected. Unlike errors stemming from a lack of medical knowledge, this failure is the inability to correctly interpret contextual qualifiers embedded in dementia care, particularly those related to disengagement and emotion regulation. When demonstrating the understanding of dementia-related perceptual limitations, many LLMs emphasized cognitive characteristics more than care principles. As a result, disengagement cues were framed as attention deficits, leading to recommending actions that may increase pressure or distress.

%% file: Content/5-Conclusion.tex
\section{Conclusion, Discussion, and Future Work}
\label{sec:conclusion}

We presented ADRD-Bench, the first benchmark dedicated to ADRD. By curating 1,438 ADRD-focused questions from seven existing medical QA datasets and introducing 149 daily caregiving questions from national, widely adopted caregiver education materials, this work provided an evaluation platform for both factual clinical knowledge and practical caregiving reasoning. The evaluation of SOTA LLMs on ADRD-Bench yielded several key findings. Different categories of models at different parameter scales presented quite different performances. While top models achieved high accuracies on questions (best performance was 0.9298 on ADRD Unified QA and 0.9597 on ADRD Caregiving QA, respectively), failure modes beyond numerical accuracy were discovered. For example, LLMs often overgeneralize clinical facts and sometimes fail to identify the major goal in a question~\cite{kim_limitations_2025}. Therefore, high overall accuracy does not guarantee true competency in ADRD care~\cite{kim_limitations_2025}. Even advanced LLMs may produce misguided responses due to shallow reasoning or misalignment with care goals~\cite{umucu_empathy_2025}. Our benchmark and evaluation together laid a groundwork for rigorously assessing LLMs in a high-impact, yet underdeveloped area of medical AI~\cite{steijger_use_2025}.

To ensure safer and more effective AI for ADRD care, domain-specific alignment techniques that bridge the gap between clinical knowledge and caregiving practices should be a focus in future research~\cite{umucu_empathy_2025}. This benchmark and our findings reinforce the importance of rigorous, targeted evaluation in high-stakes domains. As with any other medical specialty, ADRD care is a complex and evolving field~\cite{alz_association_mapping_2024}. As a result, benchmark questions and answers need to be updated in a timely manner according to the state-of-the-art practice and research. This will be a significant effort that requires the participation of the broader community (e.g., technological personnel, clinical professionals, caregiving workforce, facilities, informal caregivers, and patients) to advance safe, reliable, and compassionate AI in the ADRD domain. 

%% file: Content/Limitations.tex
\section*{Limitations}
\label{sec:limit}

As the first and a preliminary benchmark, this work has several limitations. Since the ADRD Unified QA portion was drawn from existing public benchmarks, there was a possibility that some context had been used in training certain models. Such exposure could artificially boost performance scores. Unfortunately, only a small number of models, such as medgemma-4b-it~\citep{sellergren2025medgemma} and Lingshu~\citep{xu_lingshu_2025}, explicitly reported the inclusion of some of the evaluated benchmarks in their training or fine-tuning data. The remaining models did not disclose their exact training/fine-tuning datasets related to ADRD, and thus, this information could not be traced. Current questions of ADRD Caregiving QA are in multiple-choice or true/false format, which forces models to choose an answer even when they might be uncertain. This may alter performance scores by educated guesses or the exploitation of answer cues~\cite{liu_beyond_2025}. We did not include open-ended generative tasks, since automatic evaluation of long-form responses remains challenging~\cite{singhal_large_2023}. As a result, this benchmark primarily emphasizes factual correctness and does not measure a model's ability in conversational interactions, such as proactively demonstrating empathy.

The current questions are single-turn Q\&A pairs. Multi-turn dialogues and interactive problem-solving are not included yet to test a model's ability in carrying out extended conversations, such as counseling a caregiver through a stressful situation over several exchanges, an important aspect of real-world AI assistance. In addition, caregiving decisions are often context-dependent and may vary across patient needs and professional perspectives. Although it has disclosed large disparities and challenges across SOTA models, the ADRD-Bench covers a small scope of ADRD knowledge, and it should be expanded by the broader community in future iterations for better coverage and precision.

Finally, the benchmark is developed in English and rooted in the ABC model, primarily focusing on ADRD care in the U.S. Therefore, the generalizability of ADRD-Bench to other languages and healthcare systems needs further investigation. Note that achieving high accuracy on this benchmark does not guarantee an LLM will be reliable or effective in real clinical or home-care deployment. The benchmark is a step toward identifying strengths and weaknesses. Users should be cautious not to over-interpret the results, but use them in accordance with other necessary evaluations (e.g., user studies).

%% file: Content/Ethical_Considerations.tex
\section*{Ethical Considerations}

ADRD-Bench does not involve confidential data. All content was derived from publicly available, de-identified resources. The ADRD Caregiving QA items were developed from widely used, clinically validated educational resources and reviewed by a clinician with national expertise in comprehensive ADRD care to ensure quality and relevance. This benchmark is intended for research and evaluation purposes. We caution against the direct use of LLMs' outputs in real-world clinical or caregiving settings without appropriate human oversight. Responsible use and safety should remain central considerations in downstream applications.

%% file: Content/Appendix.tex
\section{Statistics of ADRD-Related Questions}
\begin{table}[H]
\centering
\small
\caption{Number of total and ADRD-related questions in commonly used public healthcare benchmarks.}
\label{tab:adrd-question-statistics}
\begin{tabularx}{\columnwidth}{
    @{}
    >{\raggedright\arraybackslash}X
    >{\raggedleft\arraybackslash}p{0.23\columnwidth}
    >{\raggedleft\arraybackslash}p{0.30\columnwidth}
    @{}
}
\hline
\textbf{Benchmark Name} & \textbf{Number of Questions} & \textbf{Number of ADRD-Related Questions} \\
\hline
\addlinespace
\multicolumn{3}{l}{\textbf{Non-Generation Tasks}} \\
\addlinespace
\hline
MEDCALC-BENCH & 11,000  & 121 (1.10\%) \\
MEDEC        & 3,360   & 37 (1.10\%) \\
HEAD-QA       & 6,765   & 43 (0.64\%) \\
MedBullets    & 616    & 18 (2.92\%) \\
MedQA         & 12,723  & 119 (0.94\%) \\
MedMCQA      & 193,155 & 1,193 (0.62\%) \\
PubMedQA      & 1,000   & 12 (1.20\%) \\
MedHallu      & 1,000   & 16 (1.60\%) \\
\textbf{Total} & \textbf{229,619} & \textbf{1,559 (0.68\%)} \\
\hline
\addlinespace
\multicolumn{3}{l}{\textbf{Generation Tasks}} \\
\addlinespace
\hline
HealthBench   & 5,000   & 60 (1.20\%) \\
MedicationQA  & 690    & 3 (0.43\%) \\
MedDialog     & 257,469 & 945 (0.37\%) \\
MEDIQA        & 383    & 51 (13.32\%) \\
\textbf{Total} & \textbf{263,542} & \textbf{1,059 (0.40\%)} \\
\hline
\end{tabularx}

\vspace{2pt}
\parbox{\columnwidth}{\scriptsize
Sources: MEDCALC-BENCH~\citep{khandekar_medcalc-bench_2024};
MEDEC~\citep{ben_abacha_medec_2025};
HEAD-QA~\citep{vilares_head-qa_2019};
MedBullets~\citep{chen_benchmarking_2025};
MedQA~\citep{jin_what_2021};
MedMCQA~\citep{pal_medmcqa_2022};
PubMedQA~\citep{jin_pubmedqa_2019};
MedHallu~\citep{pandit_medhallu_2025};
HealthBench~\citep{arora_healthbench_2025};
MedicationQA~\citep{abacha_bridging_2019};
MedDialog~\citep{he_meddialog_2020};
and MEDIQA~\citep{abacha_overview_2019}.
}

\end{table}

\section{Preliminary Care-Context-Oriented Categorization of ADRD Unified QA}
\label{unified_qa_taxonomy}

To further characterize the content coverage of the consolidated ADRD Unified QA set, we conducted a preliminary care-context-oriented categorization of all 1,438 questions. While the questions included in ADRD Unified QA were extracted from established medical benchmarks, they were originally scattered across multiple datasets and were not explicitly organized according to their relevance to different aspects of ADRD care. Therefore, this analysis was intended to provide a descriptive characterization of the care contexts represented in existing ADRD-related benchmark content.

\subsection{Categorization Definition}

We defined five categories as below:

\begin{enumerate}
    \item \textbf{Caregiving}: Paid or unpaid care provided by family members, or care delivered by non-nurse, non-medical  licensed caregivers (e.g., community health workers, certified nurse aides)
    
    \item \textbf{Clinical Nursing}: Care provided by professional nurses (e.g., registered nurses).
    
    \item \textbf{Memory Care}: Clinical care provided by memory specialists, including physicians, physician assistants, or nurse practitioners specializing in memory clinics.
    
    \item \textbf{General Non-Memory Medical or Surgical Care}: Clinical care provided by physicians, physician assistants, or advanced practice nurses in family medicine, medicine, surgery, dermatology, pathology, or other non memory care specialty.
    
    \item \textbf{Others}: Questions scattered across other domains that are not directly related to clinical care or ADRD caregiving, but about sociology, basic science, history, epidemiology, public health, policy, and health services research, etc.
\end{enumerate}

\subsection{LLM-Assisted Categorization}

Comprehensive manual annotation of all 1,438 questions would require substantial expert effort. We therefore used GPT-5.4 to conduct an initial LLM-assisted categorization of ADRD Unified QA. Each question was assigned to exactly one of the five categories defined above. The complete prompt used for categorization is provided in Appendix~\ref{taxonomy_prompt}. This categorization was used as a descriptive analysis of the content distribution of ADRD Unified QA and was not intended to constitute a fully expert-annotated taxonomy. The results are shown in Appendix Table~\ref{tab:taxonomy_verification}.

To examine the reliability of the LLM-assisted categorization, an ADRD clinical expert manually reviewed a randomly sampled subset of 43 questions. Table~\ref{tab:taxonomy_verification} reports category-specific agreement between the LLM-assigned labels and the expert verification results. Since "Caregiving" and "Clinical Nursing" contain too few questions, the accuracies were not statistically meaningful. "Others" included many non-ADRD-specific domains, with each containing a very small number of questions. Thus, it is naturally hard to categorize due to scattered and sparse information distribution.

\begin{table}[H]
\centering
\small
\begin{tabularx}{\columnwidth}{@{}X cc@{}}
\toprule
\textbf{Category} 
& \textbf{Total Questions} 
& \textbf{Agreement} \\
\midrule
Caregiving 
& 6 
& 100\% \\

Clinical Nursing 
& 4 
& 0\% \\

Memory Care 
& 468 
& 100\% \\

\makecell[l]{General Non-Memory \\Medical or Surgical Care}
& 502 
& 90\% \\

Others 
& 458 
& 40\% \\
\midrule
\textbf{Total} 
& \textbf{1,438} 
& -- \\
\bottomrule
\end{tabularx}
\caption{Expert verification of the preliminary LLM-assisted categorization.}
\label{tab:taxonomy_verification}
\end{table}

\subsection{Categorization Prompt}
\label{taxonomy_prompt}

The following prompt was used for the preliminary LLM-assisted categorization of ADRD Unified QA. Each question was processed independently.

\begin{small}
\begin{Verbatim}[breaklines=true,breakanywhere=true,fontsize=\small,breaksymbolleft={}]
You are an expert annotator tasked with classifying ADRD-related questions into predefined categories.

Classify each question into ONE of the following categories, based strictly on the definitions below.

Category 1: Clinical Nursing
Care provided by professional nurses (e.g., registered nurses).

Category 2: Caregiving
Paid or unpaid care provided by family members, or care delivered by non-nurse, non-medical  licensed caregivers (e.g., community health workers, certified nurse aides).

Category 3: Memory Care
Clinical care provided by memory specialists, including physicians, physician assistants, or nurse practitioners specializing in memory clinics.

Category 4: General Non-Memory Medical or Surgical Care
Clinical care provided by physicians, physician assistants, or advanced practice nurses in family medicine, medicine, surgery, dermatology, pathology, or other non memory care specialty.

Category 5: Others
Questions are not related to categories 1-4.

Instructions:
- Assign exactly ONE category to each question.
- Base your classification on the primary focus of the question.
- If a question mentions multiple elements, choose the category that best represents the main intent.
- Do NOT create new categories.
- Only output one number between 1-5 related to category.
- Do not contain any explanation
- Do not contain any reasoning process

Question:
<<<
Question: {question}
A. {options[0]}
B. {options[1]}
C. {options[2]}
D. {options[3]}
>>>

Output format (only one number)
<number>
\end{Verbatim}
\end{small}

\section{Detail of Aging Brain Care (ABC) Program}
\label{abc_program}
Supported initially by the U.S. Agency for Healthcare Research and Quality, the ABC model was designed as a longitudinal, biopsychosocial intervention to address the complex clinical, behavioral, and caregiver challenges associated with ADRD. In 2006, the program demonstrated efficacy in a randomized controlled trial, showing significant reductions in behavioral and psychological symptoms of dementia, caregiver burden, and inappropriate medication use among underserved populations with high social frailty~\cite{callahan_effectiveness_2006}. Since then, the ABC program has been continuously operated, refined, and scaled, serving more than 5,000 patient–caregiver dyads~\cite{boustani2025effectiveness,lamantia_aging_2015, french_healthy_2014}. Recognized by the National Academies of Medicine, ABC integrates interdisciplinary team-based care, non-pharmacologic interventions, caregiver support, medication optimization, care coordination, and community partnerships across the full ADRD trajectory. ABC directly informed the development of an alternative payment approach for comprehensive dementia care, later adopted by the U.S. Centers for Medicare \& Medicaid Services through the GUIDE Model~\cite{haggerty_guide_2025, boustani_alternative_2019}. As such, the ABC program represents a mature, rigorously evaluated, and policy-relevant standard for high-quality ADRD care.

\section{Detail of the Inference Hardware}
\label{hardware}
For open-weight models with 14.8B parameters or fewer, all experiments were conducted on a local workstation with an NVIDIA RTX 6000 Ada Generation GPU (48 GB VRAM). For larger open-weight models with 32B parameters and above, experiments are conducted on an Oracle Cloud Infrastructure server with eight NVIDIA H100 GPUs (80 GB HBM3 memory each) or four individual NVIDIA H200 GPUs. All closed-source general models were evaluated via their official APIs/terminals.

\section{Qualitative Error Taxonomy Analysis on ADRD Caregiving QA}
\label{app:error_taxonomy}

\begin{table}[H]
\centering
\small
\begin{tabularx}{\columnwidth}{@{}X rr@{}}
\toprule
\textbf{Error Type} & \textbf{Count} & \textbf{Percentage} \\
\midrule
\makecell[l]{Misinformation of \\scientific ADRD facts} & 21 & 24.42\% \\

\makecell[l]{Misunderstanding caring/medical \\priority/focus} & 18 & 20.93\% \\

\makecell[l]{Fail to reason within \\the scope of ADRD} 
& 13 & 15.12\% \\

Over-specificity & 10 & 11.63\% \\

Sweeping statement & 9 & 10.47\% \\

\makecell[l]{Incorrect general medical \\knowledge} & 6 & 6.98\% \\

Contradictory reasoning & 5 & 5.81\% \\

Fail to recognize caring reality & 3 & 3.49\% \\

\makecell[l]{Wrong reasoning on \\a non-medical, basic truth} & 1 & 1.16\% \\
\midrule
\textbf{Total} & \textbf{86} & \textbf{100.00\%} \\
\bottomrule
\end{tabularx}

\vspace{2pt}
\parbox{\columnwidth}{\footnotesize
A total of 86 incorrect answers were generated by the two well performed models in each category. Open-weight general models: Llama-3.1-70B-Instruct, Qwen3-235B-A22B-Instruct-2507; open-weight medical models: Hulu-Med-32B, Lingshu-32B; closed-source general models: gpt-5.2-2025-12-11, claude-opus-4-5-20251101.
}

\caption{Qualitative error taxonomy for incorrect responses on ADRD Caregiving QA.}
\label{tab:error_taxonomy_distribution}
\end{table}

\section{ADRD-Bench Prompts}
\subsection{Prompt for PubMedQA}
\begin{small}
\begin{Verbatim}[breaklines=true,breakanywhere=true,fontsize=\small,breaksymbolleft={}]
You are a highly knowledgeable AI assistant specializing in biomedical sciences. Answer A for yes, B for no, or C for maybe. Do not include any explanation or additional text. Output only the letter on a single line.

Context: {context}

Question: {question}
A. yes
B. no
C. maybe

ONLY OUTPUT ONE LETTER.
Your answer MUST be exactly one of: A, B, or C.
Do NOT include explanation or text after the letter.

Your answer is:
\end{Verbatim}
\end{small}

\subsection{Prompt for HEAD-QA (Four Options)}
\begin{small}
\begin{Verbatim}[breaklines=true,breakanywhere=true,fontsize=\small,breaksymbolleft={}]
You are a highly knowledgeable AI assistant specializing in biomedical sciences. Your task is to answer multiple-choice question accurately based on the options provided. Each question will relate to biomedical concepts, and you will be asked to choose the most appropriate answer.

Select the correct answer by outputting only the letter corresponding to your choice (A, B, C or D). Please only use capital letter, without including any content and explanation.

Question: {question}
A. {options[0]}
B. {options[1]}
C. {options[2]}
D. {options[3]}

ONLY OUTPUT ONE LETTER.
Your answer MUST be exactly one of: A, B, C or D.
Do NOT include explanation or text after the letter.

Your answer is:
\end{Verbatim}
\end{small}

\subsection{Prompt for MedBullets (Four Options)}
\begin{small}
\begin{Verbatim}[breaklines=true,breakanywhere=true,fontsize=\small,breaksymbolleft={}]
You are a highly knowledgeable AI assistant specializing in medicine. Your task is to answer medical questions similar to those found on the USMLE Step 2/3 exams. You will be provided with a clinical scenario followed by several multiple-choice options.

Select the correct answer by outputting only the letter corresponding to your choice (A, B, C, or D). Please only use capital letter, without including any content. 

Clinical Scenario: {question}
A. {options[0]}
B. {options[1]}
C. {options[2]}
D. {options[3]}

ONLY OUTPUT ONE LETTER.
Your answer MUST be exactly one of: A, B, C, or D.
Do NOT include explanation or text after the letter.

Your answer is:
\end{Verbatim}
\end{small}

\subsection{Prompt for MedMCQA and MedQA}
\begin{small}
\begin{Verbatim}[breaklines=true,breakanywhere=true,fontsize=\small,breaksymbolleft={}]
You are a highly knowledgeable AI assistant specializing in medicine. Give a letter answer among A, B, C, or D. Do not include any explanation or additional text.

Question: {question}
A. {options[0]}
B. {options[1]}
C. {options[2]}
D. {options[3]}

ONLY OUTPUT ONE LETTER.
Your answer MUST be exactly one of: A, B, C or D.
Do NOT include explanation or text after the letter.

Your answer is:
\end{Verbatim}
\end{small}

\subsection{Prompt for MedHallu}
\begin{small}
\begin{Verbatim}[breaklines=true,breakanywhere=true,fontsize=\small,breaksymbolleft={}]
You are an AI assistant with extensive knowledge in the field of medicine. I want you to act as an answer judge. Given a question and an answer, your objective is to determine if the provided answer contains non-factual or hallucinated information. You SHOULD give your judgement based on the following hallucination types and the world knowledge.

You are trying to determine if there is a factual contradiction between the answer and the world knowledge. Some information in the answer might be fabricated. Answer '0' if the answer is factual and '1' if the answer is hallucinated. You can categorize the answer as hallucinated if it falls in any of the category:
Type: Misinterpretation of #Question#: These are hallucinated answers that misunderstands the question, leading to an off-topic or irrelevant response.
Type: Incomplete Information: These are hallucinated answers that Point out what is not true without providing correct information.
Type: Mechanism and Pathway Misattribution - These are hallucinated answer that falsely attribution of biological mechanisms, molecular pathways, or disease processes that contradicts established medical knowledge.
Type: Methodological and Evidence Fabrication - Inventing false research methods, statistical data, or specific clinical outcomes.

Do not return anything else, just the answer.
Return just an integer value, '0' if the answer is factual and '1' if the answer is hallucinated. No letter or word, just the integer value.

World Knowledge: {knowledge}

Question: {question}

Answer: {hallucinated answer}

ONLY OUTPUT ONE NUMBER.
Your answer MUST be exactly one of: 0 or 1.
Do NOT include explanation or text after the number.

Your answer is:
\end{Verbatim}
\end{small}

\subsection{Prompt for MEDEC}
\begin{small}
\begin{Verbatim}[breaklines=true,breakanywhere=true,fontsize=\small,breaksymbolleft={}]
The following is a medical narrative about a patient. You are a skilled medical doctor reviewing the clinical text. 

The text is either correct or contains only one error. The text has a sentence per line. Each line starts with the sentence ID, followed by a space character then the sentence to check. You should carefully check every sentence of the text. 

If every sentence in the text is correct then only return the following output: -1. 
If one sentence in the text has a medical error, only return the sentence ID of the sentence containing the error.

Make sure not to add any other words or sentences.

Clinical Note: {sentences}

ONLY OUTPUT ONE NUMBER.
Your answer MUST be exactly one number.
Do NOT include explanation or text after the number.

Your answer is:
\end{Verbatim}
\end{small}

\subsection{Prompt for ADRD Caregiving Multiple-Choice QA}
\begin{small}
\begin{Verbatim}[breaklines=true,breakanywhere=true,fontsize=\small,breaksymbolleft={}]
You are a highly knowledgeable AI assistant specializing in medicine domain. Give a letter answer among A, B, C, D, or E. Do not include any explanation or additional text.

Question: {question}
A. {options[0]}
B. {options[1]}
C. {options[2]}
D. {options[3]}
E. {options[4]}

ONLY OUTPUT ONE LETTER.
Your answer MUST be exactly one of: A, B, C, D, or E.
Do NOT include explanation or text after the letter.

Answer:
\end{Verbatim}
\end{small}

\subsection{Prompt for ADRD Caregiving True/False QA}
\begin{small}
\begin{Verbatim}[breaklines=true,breakanywhere=true,fontsize=\small,breaksymbolleft={}]
You are a highly knowledgeable AI assistant specializing in caring for patients living with Alzheimer's disease. Answer A for yes, B for no. Do not include any explanation or additional text. Output only the letter on a single line.

Question: {question}
A. yes
B. no

ONLY OUTPUT ONE LETTER.
Your answer MUST be exactly one of: A or B.
Do NOT include explanation or text after the letter.

Answer:
\end{Verbatim}
\end{small}

\section{Zero-Shot Chain-of-Thought prompts for Case Analysis}
\subsection{Prompt for ADRD Caregiving True/False QA}
\label{prompt_tf_cot}
\begin{small}
\begin{Verbatim}[breaklines=true,breakanywhere=true,fontsize=\small,breaksymbolleft={}]
You are a highly knowledgeable AI assistant specializing in caring for patients living with Alzheimer's disease. Answer A for yes, B for no.

Reason through the question step by step and explicitly write out your reasoning before giving the final answer.

Question: {question}
A. yes
B. no

First, output the reasoning process.
Then, on a new line, output the final answer as a single letter (A or B).

Format:
Reasoning:
<your reasoning>

Answer:
<ans><one letter></ans>

Reasoning:
\end{Verbatim}
\end{small}

\subsection{Prompt for ADRD Caregiving Multiple-Choice QA}
\label{prompt_mc_cot}
\begin{small}
\begin{Verbatim}[breaklines=true,breakanywhere=true,fontsize=\small,breaksymbolleft={}]
You are a highly knowledgeable AI assistant specializing in medicine. Give a letter answer among A, B, C, D, or E.

Reason through the question step by step and explicitly write out your reasoning before giving the final answer.

Question: {question}
A. {options[0]}
B. {options[1]}
C. {options[2]}
D. {options[3]}
E. {options[4]}

First, output the reasoning process.
Then, on a new line, output the final answer as a single letter (A, B, C, D, or E).

Format:
Reasoning:
<your reasoning>

Answer:
<ans><one letter></ans>

Reasoning:
\end{Verbatim}
\end{small}

\section{Completed Reasoning Output of Case Analysis Example}
\subsection{Reasoning Output for True/False Case}
\label{reasoning_tf}
Reasoning by Meta-Llama-3-8B-Instruct:
\begin{small}
\begin{Verbatim}[breaklines=true,breakanywhere=true,fontsize=\small,breaksymbolleft={}]
Reasoning: The question states that the Alzheimer's patient is unable to perform basic activities of daily living. This implies that the patient has difficulty with tasks such as bathing, dressing, eating, and using the bathroom, among others. These are essential skills for independent living, and if someone is unable to perform them, it suggests a significant impairment in their cognitive or physical abilities.

Answer: <ans>A</ans>
\end{Verbatim}
\end{small}
Reasoning by claude-opus-4-7:
\begin{small}
\begin{Verbatim}[breaklines=true,breakanywhere=true,fontsize=\small,breaksymbolleft={}]
Reasoning: Alzheimer's disease is a progressive neurodegenerative disorder. In the early stages, patients can typically still perform basic activities of daily living (ADLs) such as bathing, dressing, eating, and toileting, though they may have difficulty with more complex instrumental activities. However, as the disease progresses to moderate and severe stages, patients become increasingly unable to perform basic ADLs and require assistance. The statement is a general one without specifying the stage. In clinical contexts, Alzheimer's patients—especially as the disease advances—are characteristically known to lose the ability to perform basic ADLs, which is a hallmark feature used in staging the disease. Taking the statement as a general clinical characterization of Alzheimer's disease, it is considered true.

Answer: <ans>A</ans>
\end{Verbatim}
\end{small}

\subsection{Reasoning Output for Multiple-Choice Case}
\label{reasoning_mc}
Reasoning by qwen3-max-2025-09-23:
\begin{small}
\begin{Verbatim}[breaklines=true,breakanywhere=true,fontsize=\small,breaksymbolleft={}]
Reasoning:

When caring for someone with dementia, maintaining effective communication is crucial. Individuals with dementia often experience difficulties with attention, perception, and processing sensory information. Losing eye contact may indicate that they are distracted, overwhelmed, or simply have shifted their visual focus due to cognitive changes—not necessarily disinterest or defiance.

Option A suggests moving closer and increasing body language, which could be perceived as intrusive or overwhelming, especially if the person is already feeling overstimulated.

Option B involves following the patient’s gaze and repositioning to stay in their visual field. This is a respectful, non-confrontational way to re-establish connection without demanding attention. It aligns with best practices in dementia care, which emphasize adapting to the person’s current state rather than forcing engagement.

Option C assumes the person needs space, but losing eye contact doesn’t always mean they need a break—it might just reflect difficulty maintaining focus. Pausing unnecessarily could disrupt meaningful interaction.

Option D calling their name firmly and reminding them of the topic—can come across as scolding or pressuring, which may increase anxiety or confusion.

Option E assumes the issue is auditory distraction, but the problem described is loss of eye contact (a visual cue), not lack of hearing or response to speech.

Therefore, the most appropriate action is to gently reposition yourself within their visual field to support their ability to re-engage naturally.

Answer: <ans>B</ans>
\end{Verbatim}
\end{small}
Reasoning by gpt-5.5-2026-04-23:
\begin{small}
\begin{Verbatim}[breaklines=true,breakanywhere=true,fontsize=\small,breaksymbolleft={}]
Reasoning: In dementia care, maintaining visual engagement helps the person understand and stay oriented during conversation. If eye contact is lost, the best response is not to speak louder, be firm, or assume they need space. Instead, you should gently adjust your position so you remain within their visual field, especially because people with dementia may have reduced attention or altered visual awareness. Following their gaze and repositioning yourself helps maintain connection without startling or pressuring them.

Answer: <ans>B</ans>
\end{Verbatim}
\end{small}